\documentclass[letterpaper]{article} 
\usepackage{aaai24}  
\usepackage{times}  
\usepackage{helvet}  
\usepackage{courier}  
\usepackage[hyphens]{url}  
\usepackage{graphicx} 
\usepackage{natbib}  
\usepackage{caption} 
\usepackage{algorithm}
\usepackage{algorithmic}

\usepackage{newfloat}
\usepackage{listings}
\DeclareCaptionStyle{ruled}{labelfont=normalfont,labelsep=colon,strut=off} 
\floatstyle{ruled}
\newfloat{listing}{tb}{lst}{}
\floatname{listing}{Listing}
\usepackage{xcolor}
\usepackage{amsmath}

\title{Evaluating Theory of (an uncertain) Mind: \\ Predicting the Uncertain Beliefs of Others in Conversation Forecasting}
\author{
    Anthony Sicilia \qquad Malihe Alikhani
}
\affiliations{
    Khoury College of Computer Sciences \\
    Northeastern University \\
    \texttt{sicilia.a@northeastern.edu}
}

\usepackage{amsthm}
\usepackage{amsmath}
\usepackage{amssymb}
\usepackage{lipsum}
\usepackage{csquotes}

\usepackage{pifont}
\newcommand{\xmark}{\ding{55}}%
\usepackage{enumitem}
\usepackage{booktabs}

\begin{document}

\maketitle

\begin{abstract}
Typically, when evaluating Theory of Mind, we consider the beliefs of others to be binary: held or not held. But what if someone is unsure about their own beliefs? How can we quantify this uncertainty? We propose a new suite of tasks, challenging language models (LMs) to model the uncertainty of others in dialogue. We design these tasks around conversation forecasting, wherein an agent forecasts an unobserved outcome to a conversation. Uniquely, we view interlocutors themselves as forecasters, asking an LM to predict the uncertainty of the interlocutors (a probability). We experiment with re-scaling methods, variance reduction strategies, and demographic context, for this regression task, conducting experiments on three dialogue corpora (social, negotiation, task-oriented) with eight LMs. While LMs can explain up to 7\% variance in the uncertainty of others, we highlight the difficulty of the tasks and room for future work, especially in practical applications, like anticipating ``false uncertainty.''
\end{abstract}

\section{Introduction}
\label{sec:intro}
\begin{figure}[t]
    \centering
    \includegraphics[width=\columnwidth]{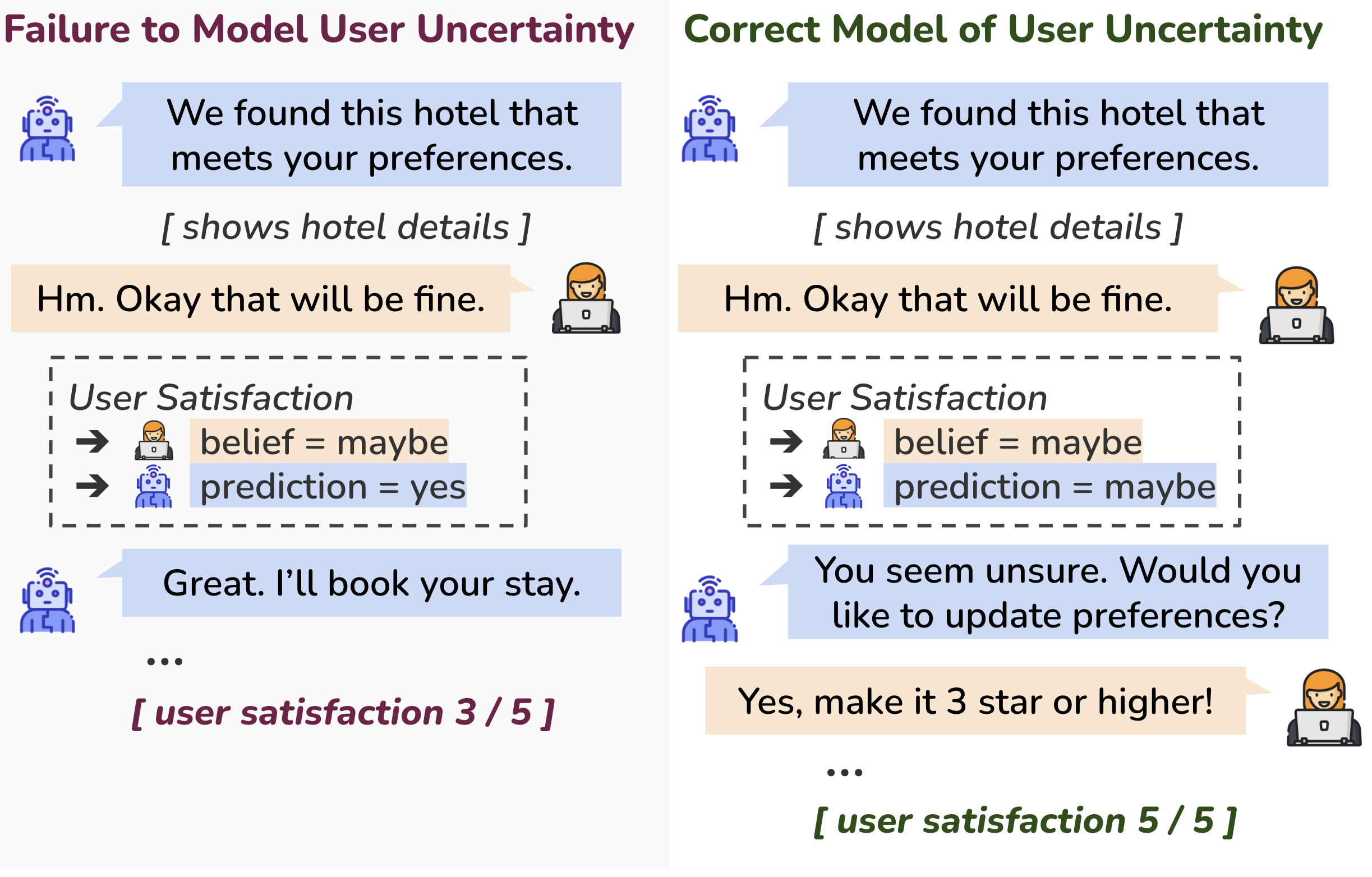}
    \caption{Recognizing uncertainty in others can influence AI dialogue strategies, ultimately improving task-success. Here, an AI assistant recognizes user uncertainty and probes to resolve it, eventually increasing user satisfaction. We formalize tasks to assess language models for their ability to recognize uncertainty in other's beliefs -- an aspect of Theory of Mind. We propose new methods to use language models for this regression task, evaluating eight models across three dialogue corpora.}
    \label{fig:intro}
\end{figure}
Theory-of-mind and, specifically, false belief prediction is an integral part of planning and decision-making in conversations \citep{ho2022planning}. For instance, an interlocutor recognizing another's false beliefs can offer clarification or explanation. While beliefs are often treated as existing in a binary state (held or not held), there are situations where an interlocutor’s belief is better represented more flexibly (e.g., held, not held, or \textit{unsure}), capturing the uncertainty or more general intensity of the belief. Strangers in a first encounter may be unsure of how they feel about each other and competing stakeholders in a negotiation may be similarly perplexed about each other’s objectives. In task-oriented settings, people may even be unsure of their goals \citep{sicilia2023isabel}, impeding success when this is not considered (e.g., see Figure~\ref{fig:intro}). How we respond in conversation depends not only on our anticipation of another's belief, but also how strongly they hold that belief. Yet, it’s unclear how conversational AI, like language models, represent the uncertainty of others in dialogue.

This paper tackles this issue, proposing new tasks to evaluate ToM capacity in language models, specifically pertaining to uncertainty. To do this, we use conversation forecasting as a tool. Whereas existing forecasting tasks \citep{sokolova-etal-2008-telling, zhang-etal-2018-conversations, sicilia2024deal}, focus on predicting aleatoric factors of uncertainty, which are inherent to the data and independent of perspective, we focus on predicting subjective and individual factors of uncertainty, held by each interlocutor. 

In particular, we ask models to forecast \textit{the uncertainty of others' beliefs} as well as \textit{the uncertainty of others' beliefs about others}.
Conversational agents capable of measuring an interlocutors beliefs (or false beliefs) can directly use this ToM capability to inform dialogue strategy. Indeed, in human-human interactions, the accuracy of interlocutor ToM directly influences dialogue behaviors \citep{bara-etal-2021-mindcraft}. For many types of beliefs, like emotions, uncertainty is an important correlate of accuracy \citep{troiano2021emotion}, providing a basis for intervening dialogue acts that can subsequently improve ToM and eventual task-success \cite{nilsenova2001uncertainty}. Recall, Figure~\ref{fig:intro} shows an example of this.

To formalize belief uncertainty, we build on a traditional statistical view, where uncertainty is measured via probability \citep{brocker2009reliability}, granting us an established framework to design our tasks. Belief uncertainty can be measured on a ternary scale (yes, no, maybe) or more general spectrum (e.g., a Likert scale) and we discuss some strategies to calibrate these human annotations to real-world probabilities. Meanwhile, to capture differences in interlocutor perspective, we disentangle this probabilistic notion into two components -- the epistemic (subjective) and the aleatoric (ground-truth) -- common to modern studies of uncertainty in machine learning \citep{hullermeier2021aleatoric}. Interestingly, our formal setup defines a series of regression tasks, allowing us to explore the relatively unexplored area of continuous inference with language models \citep{vacareanu2024words}. In this context, we study the interaction between traditional methodologies (e.g., bagging) and recent advances (e.g., self-consistent chain-of-thought, \citealp{wangself}) for the first time. 

In initiating this study of uncertainty-aware ToM evaluation in language models, we offer a few contributions:
\begin{enumerate}[nolistsep]
    \item we formalize ``false uncertainty'' -- a concept akin to false belief -- and connect it to Theory of Mind and forecasting (\S~\ref{sec:tom}), using this to motivate a new task suite (\S~\ref{sec:task_desc});
    \item we propose new methods to forecast others' uncertainty with language models (\S~\ref{sec:meth}), studying ways to use language models for regression and calibrate probability estimates with continuous labels (rather than discrete);
    \item we study how interlocutor demographics, goals, and other conversation context influences the ability of a language model at uncertainty-aware ToM (\S~\ref{sec:exp}).
\end{enumerate}
From experiments (\S~\ref{sec:exp}) across three corpora (social, negotiation, and task-oriented) and eight models, our findings suggest that, language models are able to explain some of the variance in others' uncertainty (up to 7.5\%), with the help of our methods. Yet, we also observe the difficulty of this task, even for humans. We point to areas for future work, including both applications and methodology (\S~\ref{sec:concl}), making code for tasks/evaluation open-source to promote such work.
\section{Conversation Forecasting}
\label{sec:back}
We focus on the setup of \citet{sicilia2024deal} where an agent observes a conversation and is asked to express their uncertainty about a potential outcome for this conversation; e.g., ``\textit{How much does Speaker A like Speaker B?}'' or ``\textit{Will the negotiation result in a deal?}'' As implied, the conversation is just a partial window into the true (or, eventual) ground-truth. Hidden information, such as future events or mental states, creates an inherent randomness about reality, which may not be fully determined by the available evidence. In this context, we assume a (human) agent forms a mental model capturing their uncertainty about the outcome -- a ``forecast'' about whether the outcome will occur. 
\subsection{Comparing Forecasts with Ground-Truth}
Given a (potentially partial) conversation and any accompanying evidence about the situation (e.g., interlocutor context), the forecasting agent expresses their uncertainty $P$ about the outcome of interest. For now, we assume $P$ is a probability estimate, but later allow other expressions of uncertainty (\S~\ref{sec:humans}). The forecast $P$ is evaluated by Brier score:
\begin{equation}
    \mathrm{BS} = (P - O)^2
\end{equation}
where $O$ is a binary indicator of the outcome (e.g., 1 if a deal occurs, 0 else). Forecasters with accurate uncertainty estimates (agreeing exactly with the distribution of $O$) will have a lower Brier score than other, less accurate forecasters. Brier score also ranks sub-optimal forecasts with consideration of both calibration and variance \citep{brocker2009reliability}. 
\subsection{The Missing Building Blocks for ToM}
\label{sec:tom}
We observe that Brier score, alone, does not capture the full story about an agent's uncertainty $P$. Indeed, the Brier score measures two individual aspects of uncertainty:
\begin{equation}
  \mathbf{E}[\mathrm{BS}] =  \underbrace{\textbf{Var}[O]}_{\text{aleatoric uncertainty}} + \underbrace{\mathbf{E}[(P - p)^2]}_{\text{epistemic uncertainty}}
\end{equation}
where $p$ is the probability $O=1$.
While the outcome variance captures the inherent randomness of the forecasting task, the latter quantifies the forecaster's excess errors that should not be attributed to this randomness. These model-specific aspects of error make up the \textbf{epistemic uncertainty} \citep{lahlou2022deup, hullermeier2021aleatoric}.
\paragraph{Integrating ToM in Forecasting} 
Uniquely, we consider the epistemic uncertainty of human interlocutors (treated as forecasters) in a conversation. This dual interpretation captures the individual aspects of an interlocutor's uncertainty by comparing
their forecast to ground-truth.
Precisely, it measures the fluctuations in uncertainty caused by the interlocutor themselves -- their knowledge, perceptions, and biases -- rather than those (fluctuations) which may be attributed to changes in ground-truth. Based on the epistemic uncertainty, we define an interlocutor's \textbf{false uncertainty} as:
\begin{equation}
    \mathrm{FUn} = P - p.
\end{equation}
False uncertainty similarly captures subjective fluctuations, but preserves the direction of this subjectivity, distinguishing between positive (over-confident) or negative (under-confident) forms of uncertainty. Quantifying false uncertainty will be the primary motivation for our task design. While works have focused on improving the quality of a forecast (the Brier score), ours is first to propose quantification of other interlocutors' uncertainty.
\paragraph{Other ToM Works}
ToM is often evaluated (in language models) using question-answering \citep{nematzadeh-etal-2018-evaluating, le-etal-2019-revisiting, sap-etal-2022-neural}; which, for instance, mimics common ToM evaluations from psychology, like the Sally-Anne test. Other proposals study machine ToM in situated and collaborative environments \citep{bara-etal-2021-mindcraft, ma-etal-2023-towards-holistic, li-etal-2023-theory}, focus on higher-order ToM \citep{wu-etal-2023-hi}, or consider ToM beyond (more common) belief/false belief anticipation \citep{van-duijn-etal-2023-theory}. Inference-time methods to improve ToM in language models have also been studied \citep{takmaz-etal-2023-speaking, sclar-etal-2023-minding}. Yet, whenever beliefs are involved, they are typically assumed to held or not held. Our forecasting setting is the first to consider anticipation of beliefs/false beliefs with an emphasis on the uncertainty of these beliefs.  
\paragraph{ToM Criteria} 
\citet{quesque2020theory} suggest ToM evaluations should require representation of a mental state (the other's) that differs from one's own (\textit{non-merging criterion}) and ensure task-success cannot be based on lower-level processes (\textit{mentalizing criterion}). Language models do not have mental states, so non-merging is achieved by information-asymmetry \citep{kim-etal-2023-fantom}, which forces the model to predict distinct (non-merged) perspectives. In forecasting, this distinction lies in the interlocutor uncertainty $P$ and ground-truth $p$, whenever $P \neq p$. As for \textit{mentalizing}, this criterion is undermined by spurious correlation, to which models are susceptible \citep{kim-etal-2023-fantom, shapira-etal-2024-clever}. Our current, natural conversation corpora (\S~\ref{sec:data}) do not necessarily prevent this risk. Albeit, it has been suggested socially situated tasks (like ours) can alleviate risk \citep{ma2023tomchallenges}. Using our tasks with controlled data can further lessen risk, which we leave as future work.
\section{New Uncertainty Quantification Tasks}
\label{sec:task}
\subsection{Human Expressions of Uncertainty}
\label{sec:humans}
As we are (uniquely) interested in humans as conversation forecasters, probability annotations are not necessarily the most effective way to elicit uncertainty or intensity of belief. Indeed, most of the corpora we study (\S~\ref{sec:data}) annotates belief intensity on a Likert scale; e.g., ``on a scale from 1 to 10, how much do you think Speaker B likes you.'' We focus on probability estimates because these can be compared to ground-truth world states; i.e., whether B actually ``likes,'' to enforce non-merging (\S~\ref{sec:tom}). Without ``the world'' or ``reality'' as reference, we have no way to define subjective, or false, uncertainty. Thus, we map human expressions to probability estimates to enable comparison.
\paragraph{Calibration Strategy: ``More Than Chance''}
Mapping verbal or quasi-continuous expressions of belief uncertainty to probabilities is a calibration problem; e.g., it has been approached for language models using scaling \citep{tian2023just}. In this work, we enable calibration by making a slight semantic change to the outcome of interest. Instead of studying ``whether Speaker B likes Speaker A'' we study ``whether Speaker B likes Speaker A more than would occur by chance.'' This alteration ties belief intensity annotations to a ground-truth outcome that is observable in data. Precisely, following the colloquial meaning of ``more than chance'' in the statistics literature, the ground-truth probability is defined by a $p$-value for the magnitude of the belief, computed from the data. In turn, appending ``more than chance'' defines both ground-truth outcome probabilities and an appropriate calibration function for human expressions of intensity.  We provide details in \S~\ref{sec:extreme}.
\subsection{Uncertainty Quantification (UQ) Tasks}
\label{sec:task_desc}
\paragraph{1st-Order ToM Uncertainty (1TUQ)} To quantify false uncertainty, one first needs to quantify an interlocutor's base uncertainty about their belief (i.e., the forecast $P$). Aptly, our first task evaluates a language model's ability to quantify the base uncertainty of others. For instance, suppose an interlocutor $\mathrm{A}$ expresses their uncertainty about ``whether $\mathrm{A}$ is happy'' and this is calibrated to a probability forecast $P$.\footnote{Recall, belief intensity needs calibration to a world outcome to make sense as a probability; e.g., ``A tells friends about happiness.'' We use an outcome observable in the data, i.e. ``whether A is happier than would occur by chance.''} The language model's task is to make a prediction $\hat{P}$ about the uncertainty of $\mathrm{A}$'s belief. We evaluate this prediction using regression metrics; e.g., the correlation between $P$ and $\hat{P}$, the absolute error, and the explained variance.
\paragraph{2nd-Order ToM Uncertainty (2TUQ)} Besides their own beliefs, interlocutors also hold uncertainty about the beliefs of others. 
For instance, an interlocutor $\mathrm{A}$ can express their uncertainty about ``whether interlocutor $\mathrm{C}$ likes $\mathrm{A}$''. Then, TUQ tasks the language model with quantifying the uncertainty of $\mathrm{A}$ about $\mathrm{C}$'s belief. Similar to the first-order task, we evaluate a language model's prediction by comparing it to $\mathrm{A}$'s true uncertainty.
\paragraph{False Uncertainty (FUnQ)} Finally, we ask language models to directly quantify an interlocutor's false uncertainty. In essence, this requires them to quantify both the interlocutor's uncertainty about a belief as well as the ground-truth probability that the belief is true (the outcome probability $p$). For instance, $P$ may be a forecast about ``whether interlocutor $\mathrm{C}$ likes $\mathrm{A}$'' and $p$ may be the ground-truth probability that ``$\mathrm{C}$ likes $\mathrm{A}$.'' The language model is tasked with quantifying $\mathrm{FUn} = P - p$, and we evaluate this estimation using regression metrics.
\subsection{Corpora and Basic Prompts}
\label{sec:data}
\paragraph{CaSiNo} is a corpus of negotiations about camp-resource allocation \citep{chawla2021casino}. Interlocutors barter over available resources, such as fire-wood and water, based on (assigned) resource preferences. Performance-based monetary incentives stimulate competitive behaviors. Interlocutors indicate their satisfaction with the final deal on a 5-point scale. For an interlocutor $\mathrm{A}$, we ask language models to predict ``how certain $\mathrm{A}$ is that they are more satisfied than would occur by chance.'' Precise details are in \S~\ref{sec:prompts}. This formulation allows us to evaluate language models for 1st Order ToM uncertainty quantification (1TUQ). The average number of tokens in a conversation is 320.
\paragraph{CANDOR} is a corpus of spoken conversations between strangers, conducted over video communication platform \citep{reece2023candor}. Conversations are social in nature with minimum time constraints and an assigned goal of ``getting to know each other.'' Exit interviews (conducted privately) ask interlocutors to quantify how much they like each other on a 7-point scale, as well as how much they \textit{think} their conversation partner likes them. For two interlocutors $\mathrm{A}$ and $\mathrm{B}$, we ask language models to predict ``how certain is $\mathrm{B}$ that they like $\mathrm{A}$ more than would occur by chance.'' As with CaSiNo, this lets us evaluate language models at the first-order task (1TUQ). Because of the available data, we also ask language models to predict ``how certain is $\mathrm{A}$ that $\mathrm{B}$ likes $\mathrm{A}$ more than would occur by chance.'' As we discuss in \S~\ref{sec:meth}, this lets us evaluate models at the second-order task (2TUQ) and False Uncertainty Quantification (FUnQ). The average token-count is 11K, but we only show models the first 5K.
\paragraph{MultiWOZ} is a task-oriented Wizard-of-Oz corpus wherein one human plays the role of a conversational booking system (for hotels, restaurants, etc.) and the other plays as user \citep{eric-etal-2020-multiwoz}. Additional annotations \citep{sun2021simulating} designate perceived satisfaction of the user by crowd-workers on a 5-point scale. This annotation is less organic than previous datasets, but it can be considered a representative proxy, capturing how annotators might feel if they were in the user's position. As such, we ask language models to predict ``how certain the user is that they are more satisfied than would occur by chance'' using the average crowd-worker annotation as ground-truth. This allow us to evaluate 1TUQ. The average token-count is 460.
\section{Methods}
\label{sec:meth}
\subsection{Forecasting the Uncertainty of Beliefs}
\label{sec:belief_pred}
Direct Forecasting (\texttt{DF}, \citealp{sicilia2024deal}) is a good ``out-of-the-box'' method for uncertainty-aware conversation forecasting with language models. Adapted to our belief anticipation problem, we prompt the language model to express its predicted uncertainty for the interlocutor on a 10-point scale. We parse the prediction directly from the model's sampled completion and divide by 10 to get an estimate $\hat{P}$ for the interlocutor's true forecast $P$. In general, we use the Chain of Thought (CoT) strategy proposed by \citet{kojima2022large}, asking the model to approach the prediction ``step-by-step.''
\subsubsection{Post-Hoc Scaling}
\label{sec:scaling}
Post-hoc scaling (calibration) tends to improve direct forecasts \citep{tian2023just, sicilia2024deal}, requiring only a small amount of data. 
Notably, our ToM tasks work with continuous uncertainty annotations in place of traditional, discrete outcome annotations. We propose new scaling methods to accommodate our data.
\paragraph{Platt Scaling (PS)} One option is to assume the relationship between the true uncertainty $P$ and the predicted uncertainty $\hat{P}$ is linear in the logits; e.g., this is common in soft classification \citep{platt1999probabilistic}. In our new setting,
\begin{equation}
\label{eqn:platt}
    \mathrm{logit}(P) \approx \alpha \cdot \mathrm{logit}(\hat{P}) + \beta.
\end{equation}
The new (re-scaled) forecast is:
\begin{equation}
    \hat{P}_{\mathrm{PS}} = \mathrm{expit} \big (\alpha \cdot \mathrm{logit}(\hat{P}) + \beta \big )
\end{equation}
where $\alpha$, $\beta$ are the OLS estimates of Eq.~\eqref{eqn:platt}.
\paragraph{Linear Scaling (LS)} We also suggest linear scaling, which instead learns a direct linear map:
\begin{equation}
    \hat{P}_{\mathrm{LS}} = \mathrm{clip}(\alpha \cdot \hat{P} + \beta, 0, 1).
\end{equation}
\subsubsection{Fine-Tuning a Regression Head}
Fine-tuning a classification head on a language model's latent features 
can help forecasting performance in soft classification \citep{kadavath2022language}. Again, since our annotations are continuous, we slightly modify this, replacing the classification head with a regression head. Specifically, using the same prompt as \texttt{DF}, the language model encodes latent features $\mathbf{x}$ and inference is conducted as:
\begin{equation}
    \hat{P}_{\mathrm{FT}} = f_\theta(\mathbf{x})
\end{equation}
where $f_\theta$ is the regression head. We test a linear regression head (\texttt{FT L}), a 2-layer ReLU-network head (\texttt{FT NN}), and a random forest head (\texttt{FT RF}). Details are in \S~\ref{sec:optim}.
\subsubsection{Bias and Variance}
\label{sec:b_and_v}
Classically, for a fixed probability $P$, the MSE of a corresponding estimate $\hat{P}$ can be decomposed:
\begin{equation}
    \mathbf{E}[(P - \hat{P})^2] = \textbf{Var}[\hat{P}] + \textbf{Bias}^2(P, \hat{P})
\end{equation}
where $\textbf{Bias}(P, \hat{P}) = \mathbf{E}[\hat{P}] - P$. This points to two possible ways we can reduce error, discussed next.
\paragraph{Bagging} or Bootstrap Aggregating trains many models on random samples from the same data and averages the predictions of all the models. For instance, this is how random forests (\texttt{RF}) are trained \citep{breiman2001random}. It is a known variance reduction strategy, and is why we explore \texttt{RF} as a regression head.
Another strategy we can take, with language models, is to ``bag'' the Chain-of-Thought (CoT) inferences generated when we ask a model to think ``step-by-step''; i.e., we suggest re-sampling the direct forecasts produced by this CoT prompt many times ($n=10$) and averaging these to make an inference. The changes in sampling distribution triggered by each new explanation make this distinct from greedy decoding (see Table~\ref{tab:mce_v_temp}). This strategy is akin to traditional bagging, except we re-sample model explanations, instead of training data. We call this approach a ``Bag of Thoughts'' (\textbf{BoT}). Our strategy is also similar to, yet distinct from, self-consistent decoding in classification \citep{wangself}, where a majority vote is used to aggregate multiple CoT inferences. With majority votes, variance reduction is not necessarily a plausible motivation, since the bias-variance trade-off is not well defined in classification \citep{brown2024bias} and reducing variance can actually increase errors \citep{james2003variance}. Ours is the first work to connect CoT aggregation to variance reduction. Indeed, like bagging, one can show BoT lowers the variance of the direct forecasts (i.e., compared to $n=1$) using basic properties of variance. 
\paragraph{Demographic Data} provides important background information about the interlocutor in question. Particularly, we consider the age, sex, race, and education of the interlocutor. We hypothesize these characteristics reduce prediction bias because they add situational context, an important aspect of Theory of Mind \citep{ma-etal-2023-towards-holistic}. Moreover, language models are known to inherent and propagate certain social biases \citep{gallegos2023bias} and improved context representations -- such as those achieved by making demographics clear -- can be an effective means to mitigate the biases in generative inference \citep{sicilia2023learning}.
\subsection{Forecasting False Uncertainty}
A straightforward way to predict false uncertainty is to have the model shift perspectives across two inference steps. For instance, in CANDOR, we predict $\mathrm{A}$'s uncertainty about ``whether $\mathrm{C}$ likes $\mathrm{A}$...'' and then predict the ground-truth probability that ``$\mathrm{C}$ likes $\mathrm{A}$...'', shifting perspectives from $\mathrm{A}$ to the outside world. Denoting these $\hat{P}$ and $\hat{p}$, respectively, the following estimates false uncertainty:
\begin{equation}
    \hat{\mathrm{FUn}} = \hat{P} - \hat{p}.
\end{equation}
The same strategies discussed in \S~\ref{sec:belief_pred} can be used to make the individual components of this inference, learning $\hat{p}$ and $\hat{P}$, separately. Alternatively, one can estimate $\hat{\mathrm{FUn}}$ directly. This makes the most sense in the fine-tuning setting, where both latent representations (from the $\hat{p}$ and the $\hat{P}$ prompt) can interact in a (non-linear) regression head to improve inference. We append \texttt{J} (joint learning) to denote this strategy.
\section{Experiments}
\label{sec:exp}
\begin{table}[]
\centering\small
\begin{tabular}{llllll}
   \texttt{method} & \texttt{xl}    & MAE     & $R^2$     & $\min$ & $\max$ \\\cmidrule(lr){1-1}\cmidrule(lr){2-2}\cmidrule(lr){3-3}\cmidrule(lr){4-6}
\texttt{DF}  & \xmark    & 44.9 & -370 & -620  & -110   \\
\texttt{DF (LS)}& \xmark & \textbf{22.2} & \textbf{1.5} & -4.1  & 7.5  \\
\texttt{DF (PS)}& \xmark & 31.9 & -180 & -740  & -1.8  \\
\texttt{FT (L)} & \xmark &  22.5    &  -1.0     & -2.3     & 1.1  \\\hline
\texttt{DF (LS)}& \checkmark  &   22.1    &  \textbf{2.1} & -5.3  & 12.5 \\ 
\texttt{FT (L)} &\checkmark  &  22.2    &  0.1 & -5.1  & 3.7 \\
\texttt{FT (NN)} &\checkmark  &  22.3    &  -9.2 & -30.5  & 5.4 \\   
\texttt{FT (RF)} &\checkmark  &  \textbf{21.9}    &  1.3 & -5.4  & 7.6 \\
\end{tabular}
\caption{ Regression performance on first-order Theory of Mind uncertainty quantification (1TUQ). No BoT or demographic data is used. Direct forecasts (\texttt{DF}) are linearly correlated ($R=0.14$) before scaling, but only linear post-hoc scaling (\texttt{LS}) calibrates these forecasts to be good predictors ($R^2$ up to $12.5\%$). A slight non-linear, monotone relationship also exists ($\rho = 0.16$), but this is not well-modeled by Platt scaling. Tuning a regression head with 8x more data or about 800 data points (\texttt{xl}) does not explain more variance. With the same data, scaling forecasts offers improvement.}
\label{tab:tuq-1}
\end{table}
\begin{table*}[]\centering\small
\begin{tabular}{llllllllll}
\textbf{BoT} & \textbf{DEM} & \texttt{Llama3 8B} & \texttt{Mix 8x7B} & \texttt{Gemma 7B} & \texttt{GPT 3.5} & \texttt{Avg} & \texttt{Llama3 70B} & \texttt{Mix 8x22B} & \texttt{GPT 4o} \\\cmidrule(lr){1-2}\cmidrule(lr){3-7}\cmidrule(lr){8-10}
\xmark      & \xmark           & 0.7      & 1.3        & 0.0        & -0.3  & 0.4 & 0.1       & 2.5         & 2.2  \\
\xmark      & \checkmark          & -1.5     & 0.7        & 0.7    & -1.0  & -0.3 & \textbf{0.7}       & \textbf{3.2}         & \textbf{3.3}  \\\hline
\checkmark     & \xmark           & \textbf{0.8}      & \textbf{1.4}        & \textbf{1.9}    & 0.0    &  1.0 & \xmark            &    \xmark           &     \xmark   \\
\checkmark     & \checkmark          & 0.8      & 0.6        & 0.7    & \textbf{0.6} & 0.7 &  \xmark & \xmark           &    \xmark          
\end{tabular}
\caption{ Explained variance (\%) micro-averaged across CANDOR and CaSiNo for 1TUQ task. Results ablate demographic data (in prompt) and BoT (\S~\ref{sec:b_and_v}). Demographic context (DEM) tends to only help larger model's direct forecasts (LS), while smaller models fail to make use of it. Bagging (BoT) tends to help, or have no effect, on smaller models. At least, it mitigates degradation of $R^2$ when including demographic data.}
\label{tab:tuq-1-ablation}
\end{table*}
\begin{table*}[]
\centering\small
\begin{tabular}{llllllllll}
\textbf{Dataset}  & \texttt{Llama3 8B} & \texttt{Mix 8x7B} & \texttt{Gemma 7B} & \texttt{GPT 3.5} & \texttt{Llama3 70B} & \texttt{Mix 8x22B} & \texttt{GPT 4o} & \texttt{Avg}& \texttt{Hum}  \\\cmidrule(lr){1-1}\cmidrule(lr){2-5}\cmidrule(lr){6-8}\cmidrule(lr){9-9}\cmidrule(lr){10-10}
\textbf{CANDOR}   & -0.2    & \textbf{2.3}       & 0.2   & 0.1   & -0.4      & \textbf{1.0}          & 0.6 & 0.5 & 2.7 \\
\textbf{CaSiNo}   & 1.7      & 0.5        & \textbf{3.6}   & 0.0       & 0.6       & \textbf{3.9}         & 3.7 & 2.0 & \xmark \\
\textbf{MultiWOZ} & \textbf{4.8}      & 2.6        & 4.4    & 3.3   & 4.5       & 2.8         & \textbf{6.8} & 4.2 & \xmark \\\hline
\textbf{Avg} & 2.1 (0.6) & 1.8 (1.6) & 2.7 (0.7) & 1.1 (0.0) & 1.5 & 2.5 & \textbf{3.7}
\end{tabular}
\caption{Explained variance for direct forecasts (LS) on 1TUQ task separated by data and model. BoT is used for small models. Parentheses show $R^2$ without bagging. Models show varied success on different data, meanwhile BoT improves small models to outperform others 10x larger. \texttt{Hum} denotes human $R^2$, after LS (MAE=17.7).}
\label{tab:tuq-1-datasets}
\end{table*}
\begin{table*}[]
\centering\small
\begin{tabular}{llllllllll}
\textbf{BoT} & \textbf{DEM} & \texttt{Llama 3 8B} & \texttt{Mix 8x7B} & \texttt{Gemma 7B} & \texttt{GPT 3.5} & \texttt{Avg} & \texttt{Llama 3 70B} & \texttt{Mix 8x22B} & \texttt{GPT 4o}  \\\cmidrule(lr){1-2}\cmidrule(lr){3-7}\cmidrule(lr){8-10}
\xmark  & \xmark     & -1.4       & -0.9         & -0.7     & -0.4   & -0.9 & 0.4         & 1.0             & 0.1  \\
\xmark  & \checkmark & 0.0          & 1.4          & -0.7     & -0.2  & 0.1  & \textbf{0.7}         & \textbf{1.8}           & \textbf{0.7} \\\hline
\checkmark & \xmark     & \textbf{0.8}        & 0.6          & -1.8     & \textbf{1.8}   & 0.4  &      \xmark       &         \xmark      &    \xmark   \\
\checkmark & \checkmark    & 0.6        & \textbf{1.5}          & \textbf{-0.6}     & -0.3 & 0.3  &    \xmark         &   \xmark            &    \xmark    
\end{tabular}
\caption{Explained variance (\%) for 2TUQ task on CANDOR dataset. As before, bagging (BoT) helps smaller models, sometimes pushing them to perform at the level of their larger counterparts. In second-order ToM UQ, demographic data (DEM) appears to help most models, albeit bagging complicates this result for smaller models.}
\label{tab:tuq-2-ablation}
\end{table*}
\begin{table}[]
\centering\small
\begin{tabular}{llllll}
   \texttt{method} & \texttt{xl}    & MAE     & $R^2$     & $\min R^2$ & $\max R^2$ \\\cmidrule(lr){1-1}\cmidrule(lr){2-2}\cmidrule(lr){3-3}\cmidrule(lr){4-6}
\texttt{DF (LS)}  & \xmark    & 24.2 & -0.9 & -2.5  & -0.1   \\\hline
\texttt{DF (LS)}& \checkmark  &   24.3    &  -1.9 & -7.0  & -0.2 \\ 
\texttt{FT (RF)} &\checkmark  &  24.8    &  -5.2 & -8.9  & -2.2 \\
\texttt{FT (RF-J)} &\checkmark  &  23.9    &  0.3 & -1.7  & 1.4
\end{tabular}
\caption{Regression metrics for False Uncertainty prediction (FUnQ) on CANDOR. Even with more data, neither direct forecasting (LS) nor fine-tuning is able to explain variance in the ground-truth false uncertainty. False uncertainty prediction is a more difficult task, requiring models to perspective shift and properly predict both the interlocutor's uncertainty and that of the outside world, accurately and simultaneously.}
\label{tab:funq}
\end{table}
\subsection{Setup}
\paragraph{Data \& Seeds} We use the 3 datasets/prompting schemes discussed in \S~\ref{sec:data}. More details on prompts are in \S~\ref{sec:prompts}. We use 5 different random seeds to create 5 distinct train/test splits. For training, $n=100$ unless otherwise noted.
\paragraph{Models} Direct forecasting (\texttt{DF}) is conducted with \texttt{Llama3 8B} and \texttt{70B} \citep{llama3modelcard}, \texttt{Mixtral 8x7B} and \texttt{x22B} (v0.1 \citealp{jiang2024mixtral}), \texttt{Gemma 7B} \citep{gemmateam2024gemma}, \texttt{GPT 3.5} (turbo-0125, OpenAI) and \texttt{GPT-4o} (2024-05-13, OpenAI). All models are instruction-tuned (chat) versions. We use default sampling parameters, given on the API or model repository. For fine-tuning (\texttt{FT}), we use latent representations from a pre-trained masked language model, specifically fine-tuned for long-context embedding (M2-BERT, \citealp{fu2024monarch}), which regularly beats much larger models at embedding tasks \citep{monarch}. We use Together AI and Open AI APIs for inference.  
\paragraph{Metrics} We report standard regression metrics including the Pearson (linear) correlation $R$, the Spearman (rank) correlation $\rho$, the mean absolute error $\mathrm{MAE}$, and the \% of variance in the test data explained by the predictions $R^2$. Explaining more variance is better, but it's not typical to explain all of it ($R^2 = 100\%$). For reference, 
explained variance in (traditional) forecasting tasks with language models is low (up to 10\% Brier Skill Score -- a type of explained variance for soft classifiers, \citealp{sicilia2024deal}). Generally, we use train data to compute the mean when estimating variance on the test set (called ``out-of-sample'' $R^2$ ), which provides a fair evaluation on held-out sets. In this context, $R^2$ can also be interpreted as percent improvement compared to a constant mean prediction. For $\mathrm{MAE}$, we report the \textit{additive} error in \% probability (e.g., $\lvert 0.2 - 0.4 \rvert \times 100\% = 20\%$). Finally, metrics are micro-averaged over all data splits.
\subsection{Results \& Analysis}
We structure our results using a research question (RQ) / answer (A) format, with detailed discussion following each.
\begin{displayquote}
\textit{RQ1: Can language model's predict the uncertainty of other interlocutors in a conversation?} \\
\textit{A: No. Inference ``out-of-the-box'' is poor, but some post-hoc scaling methods enable better prediction.}
\end{displayquote}
\paragraph{Comparison of Scaling Methods}
Table~\ref{tab:tuq-1} reports regression metrics for first-order ToM UQ (1TUQ) split according to different methods of inference. While direct forecasts are ineffective ``out-of-the-box,'' linear scaling (\texttt{DF} \texttt{LS}) with 100 data points can improve scores to a positive explained variance, on average. These results suggest a consistent (if slight) linear relationship between the language model's inferences and the interlocutors' true uncertainty. Explained variance sometimes exceeds 7\%, or with more data, 12\%. Contrary to conventional wisdom (using soft classifiers to forecast outcomes), a logit-linear relationship between the model's inferences and it's target seems unlikely, due to the poor performance of \texttt{DF} \texttt{PS}. 
\begin{displayquote}
\textit{RQ2: Does variance reduction via bagging improve language model capability at our regression task?} \\
\textit{A: Yes. Random forests trained on language model embeddings show promise. The proposed Bag of Thoughts (BoT) strategy also improves inference.}
\end{displayquote}
\paragraph{Variance Reduction Strategies} Use of bagging in fine-tuning (i.e., via random forests) did improve performance as anticipated, compared to other tuning strategies. We recall, bagging is a known variance reduction strategy, which can ultimately reduce errors by this mechanism. Another variance reduction strategy we propose is Bag of Thoughts (BoT). Table~\ref{tab:tuq-1-ablation} reports ablation study of BoT for first-order TUQ, limited to CANDOR and CaSiNo. Ablation is also reported for second-order TUQ, limited to CANDOR, in Table~\ref{tab:tuq-2-ablation}. Findings show BoT has positive impact on small models on average, with particular models/setups seeing substantial gain (2\% bump for \texttt{Gemma 7B} on 1TUQ, more for \texttt{GPT 3.5} on 2TUQ). Performance is amplified more so in Table~\ref{tab:tuq-1-datasets} (includes MultiWOZ). Averaged across all 1TUQ data, BoT allows small models to surpass some large models (particularly, \texttt{Llama3 70B}). We did not try BoT for large models, as their lower throughput (tokens/second) made repeated sampling time consuming. Comparison between BoT and greedy decoding is in Table~\ref{tab:mce_v_temp}.
\paragraph{Why BoT Works} Because we use BoT on direct forecasts, and \textit{then} scale them, BoT actually reduces variance in the feature space of the linear scaling function (not necessarily the predictions). Indeed, comparing before/after BoT shows an increase in the standard deviation of the prediction (+0.5\% proba.). Meanwhile, in feature space (the pre-scaled forecast), the STD decreases by 1.5\% probability. Our hypothesis is that variance reduction in feature space (by BoT) actually increases the signal-to-noise-ratio, mitigating the effects of outlier inferences from the model. Increase correlation between pre-scaled forecasts and ground-truth after applying BoT (+0.03) may confirm this.
\begin{displayquote}
\textit{RQ3: Can interlocutor demographic information be used by an LM to improve ToM UQ?} \\
\textit{A: Yes, depending on model size. Larger models show more consistent ability to use demographic context.}
\end{displayquote}
\paragraph{Use of Demographic Context}
In Table~\ref{tab:tuq-1-ablation} we ablate the role of including demographic data in the prompt (\textbf{DEM}), limited to first-order TUQ on CANDOR and CaSiNo. With or without BoT, adding demographics tends to hurt performance of small models (0.7\% and 0.4\% drop in average $R^2$, respectively). Meanwhile, the scaled inferences of larger models are all improved by including demographics. 
In similar ablation for second-order TUQ (Table~\ref{tab:tuq-2-ablation}), we did find less conclusive evidence of a distinction between smaller and larger models use of demographic context. Without BoT, demographics seem to help both small/large models, but with BoT smaller models show mixed responses. 
\paragraph{Demographics and Bias}
Our initial hypothesis was the bias reduction was the principle mechanism by which demographic context could reduce error. This is consistent (for large models, 1TUQ) with observed reduction in bias after including demographics (-0.1\%). The limited effect size does suggest potential for other factors. For instance, similar to variance reduction, interplay between demographic data and scaling may play a role. On the other hand, models (in general) may be ineffective in using demographic context.
\begin{displayquote}
\textit{RQ4: What factors of conversation context and ToM task impact LM inferences about uncertainty?} \\
\textit{A: Longer conversations are more difficult for LMs. Humans may also hide goals in some contexts (e.g., negotiation), compounding difficulty. LMs also show difficulty when shifting perspectives.}
\end{displayquote}
\paragraph{Data Comparison} 
Table~\ref{tab:tuq-1-datasets} reports explained variance for 1TUQ for \texttt{DF LS}, split by model and dataset. We observe CANDOR to be the most difficult dataset for 1TUQ, followed by CaSiNo, then MultiWOZ. One hypothesis for the difficulty of CANDOR is the length of it's conversations, which average more than 11K tokens (GPT-2 tokenizer). This may also be compounded because dialogue is between strangers. Small, but important, nuances can become dominated other -- perhaps, superficially polite -- interactions. The reality that humans can hide their true mental states may also explain increased difficulty in CaSiNo, a negotiation corpus. Rather than ``acting'' to be polite, interlocutors in the CaSiNo corpus hide motives and intentions, as a strategy, to receive a better deal. In contrast, in the collaborative and task-oriented MultiWOZ corpus, interlocutors have incentive to reveal many aspects of their mental state; e.g., to indicate satisfactory constraints for their booking task. 
\paragraph{Task Comparison} Our methods are effective for (at least some models) on both first-order and second-order TUQ. We refrain from commenting on the differences in difficulty between these, since it varies widely depending on model. The task that stood our was False Uncertainty Quantification (FUnQ), which has results split by method in Table~\ref{tab:funq}. The key observation is that, unlike previous tasks, no model achieves positive explained variance via direct forecasting, even after scaling. This is evidenced by the negative maximum $R^2$ value. The best fine-tuning strategies offer some improvement, but performance is still relatively low.
\paragraph{Why FUnQ is Hard}
We hypothesize this difficulty may, in part, come because model errors compound across multiple inference steps; e.g., the inference for interlocutor's uncertainty $\hat{P}$ and the inference for the ground-truth probability $\hat{p}$. Indeed, one data point in favor of this hypothesis is the positive explained variance of the joint fine-tuning procedure (\texttt{FT} \texttt{RF-J}), which conducts non-linear inference over the embedding of both prompts (i.e., to infer $\hat{P}$ and $\hat{p}$) and then produces a single estimate for the difference $P - p$. 
\begin{displayquote}
    \textit{RQ5: How do humans perform at ToM UQ tasks?} \\
    \textit{A: Slightly better than LMs (scaling is still needed).}
\end{displayquote}
\paragraph{Human Performance} Because of available annotations (\S~\ref{sec:data}), we infer human performance at first-order ToM UQ on CANDOR. Interestingly, linear scaling also improves the performance for human forecasts, which may be suggestive of individual baselines for how people express their uncertainty (or, intensity) about beliefs. Human performance is not drastically higher than models ($R^2=2.7\%$, MAE=17.7), which is again suggestive of the difficulty of this corpus (recall, our data comparison).
\begin{displayquote}
    \textit{RQ6: Can uncertainty estimates improve LM inference at routine ToM (i.e., belief classification)?} \\
    \textit{A: Yes. In \S~\ref{sec:case_study}, we conduct a case-study with Llama 3 and CaSiNo. Using uncertainty estimates to infer beliefs improves accuracy and F1 score.}
\end{displayquote}
\section{Conclusions}
\label{sec:concl}
This paper details tasks and methods to explore how language models represent the uncertainty of other interlocutors in conversations. We make connections between this task and Theory of Mind, suggesting a continuous analog of false beliefs (we call, false uncertainty) and asking models to quantify it. High-level findings are summarized below:
\begin{itemize}[nolistsep]
    \item language models can explain variance in others' uncertainty on some corpora (up to 7\%), but humans may hide mental states in social dialogue and negotiations;
    \item humans themselves can have trouble with this task (in social dialogue) and appear to have different baselines for the intensity of their beliefs;
    \item variance and bias reduction are important mechanisms for tailoring language models to regression tasks;
    \item perspective shift in uncertainty modeling is difficult, as language models experience compounding errors that prevent accurate prediction of false uncertainty.
\end{itemize}
We also highlight some areas of future work.
\paragraph{Dialogue Policy and Generation} A motivation of our work is machine dialogue, as recognizing false uncertainty can inform dialogue acts, like explanation, as well as how to execute such an act. Prior work uses ToM to directly inform dialogue policies and generation \citep{zhou-etal-2023-cast}, and our ToM UQ methods may be fruitful in such contexts.
\paragraph{Communication Theories} Accurate and automated quantification of uncertainty in interlocutors has the potential to scale up studies of uncertainty in communication, and communication theories founded on uncertainty \citep{berger1974some, sunnafrank1986predicted} to large corpora. For instance, in data driven studies of these theories, participant annotation of uncertainty is needed \citep{yoo2009uncertainty}, but may be relaxed with suitable automated methods. The role of uncertainty in alternate communicative theories, such as grounding \citep{clark1991grounding} is also of interest.
\paragraph{Demographics and Uncertainty} Our methodological findings -- that demographics can improve inference about others' uncertainty -- suggest these factors may play an important role in how individuals express and manage uncertainty. Further study of this connection can prevent unintended biases in the dialogue systems we create.
\paragraph{Methodology} We leave ample room for improvement on these ToM UQ tasks. While fine-tuning showed potential in our results, the scale of available data was a limiting factor. Work-arounds, e.g., simulation, can enable full model fine-tuning, as in previous forecasting work \citep{sicilia2024deal}. False uncertainty quantification, in particular, may benefit from new methodologies which can improve models' ability to shift perspectives -- a more broadly useful capability; e.g., for LLM-based agent simulations \citep{park2023generative}.
\section*{Acknowledgements} This research was supported in part by Other Transaction award HR0011249XXX from the U.S. Defense Advanced Research Projects Agency (DARPA) Friction for Accountability in Conversational Transactions (FACT) program. Thanks to Mert Inan for helpful feedback.
\clearpage
\bibliography{custom}
\section*{Reproducibility Checklist}
This paper:
\begin{itemize}
    \item Includes a conceptual outline and/or pseudocode description of AI methods introduced (\textbf{yes}/partial/no/NA)
    \item Clearly delineates statements that are opinions, hypothesis, and speculation from objective facts and results (\textbf{yes}/no)
    \item Provides well marked pedagogical references for less-familiare readers to gain background necessary to replicate the paper (\textbf{yes}/no) 
\end{itemize}

Does this paper make theoretical contributions? (yes/\textbf{no})

If yes, please complete the list below.
\begin{itemize}
    \item All assumptions and restrictions are stated clearly and formally. (yes/partial/no)
    \item All novel claims are stated formally (e.g., in theorem statements). (yes/partial/no)
    \item Proofs of all novel claims are included. (yes/partial/no)
    \item Proof sketches or intuitions are given for complex and/or novel results. (yes/partial/no)
    \item Appropriate citations to theoretical tools used are given. (yes/partial/no)
    \item All theoretical claims are demonstrated empirically to hold. (yes/partial/no/NA)
    \item All experimental code used to eliminate or disprove claims is included. (yes/no/NA)
\end{itemize}

Does this paper rely on one or more datasets? (\textbf{yes}/no)

If yes, please complete the list below.
\begin{itemize}
    \item A motivation is given for why the experiments are conducted on the selected datasets (\textbf{yes}/partial/no/NA)
    \item All novel datasets introduced in this paper are included in a data appendix. (yes/partial/no/\textbf{NA})
    \item All novel datasets introduced in this paper will be made publicly available upon publication of the paper with a license that allows free usage for research purposes. (yes/partial/no/\textbf{NA})
    \item All datasets drawn from the existing literature (potentially including authors’ own previously published work) are accompanied by appropriate citations. (\textbf{yes}/no/NA)
    \item All datasets drawn from the existing literature (potentially including authors’ own previously published work) are publicly available. (\textbf{yes}/partial/no/NA)
    \item All datasets that are not publicly available are described in detail, with explanation why publicly available alternatives are not scientifically satisficing. (yes/partial/no/\textbf{NA})
\end{itemize}

Does this paper include computational experiments? (\textbf{yes}/no)

If yes, please complete the list below.

\begin{itemize}
    \item Any code required for pre-processing data is included in the appendix. (\textbf{yes}/partial/no).
    \item All source code required for conducting and analyzing the experiments is included in a code appendix. (\textbf{yes}/partial/no)
    \item All source code required for conducting and analyzing the experiments will be made publicly available upon publication of the paper with a license that allows free usage for research purposes. (\textbf{yes}/partial/no)
    \item All source code implementing new methods have comments detailing the implementation, with references to the paper where each step comes from (\textbf{yes}/partial/no)
    \item If an algorithm depends on randomness, then the method used for setting seeds is described in a way sufficient to allow replication of results. (\textbf{yes}/partial/no/NA)
    \item This paper specifies the computing infrastructure used for running experiments (hardware and software), including GPU/CPU models; amount of memory; operating system; names and versions of relevant software libraries and frameworks. (\textbf{yes}/partial/no)
    \item This paper formally describes evaluation metrics used and explains the motivation for choosing these metrics. (\textbf{yes}/partial/no)
    \item This paper states the number of algorithm runs used to compute each reported result. (\textbf{yes}/no)
    \item Analysis of experiments goes beyond single-dimensional summaries of performance (e.g., average; median) to include measures of variation, confidence, or other distributional information. (\textbf{yes}/no)
    \item The significance of any improvement or decrease in performance is judged using appropriate statistical tests (e.g., Wilcoxon signed-rank). (yes/partial/\textbf{no})
    \item This paper lists all final (hyper-)parameters used for each model/algorithm in the paper’s experiments. (\textbf{yes}/partial/no/NA)
    \item This paper states the number and range of values tried per (hyper-) parameter during development of the paper, along with the criterion used for selecting the final parameter setting. (\textbf{yes}/partial/no/NA)
\end{itemize}

\clearpage
\appendix
\section{Appendix}
\label{sec:appendix}
\subsection{Extreme Value Uncertainty}
\label{sec:extreme}
Given an annotation $m$ about an interlocutor $\mathrm{A}$'s magnitude of belief, we consider the forecasting problem with outcome $\gamma=$ ``whether $\mathrm{A}$'s magnitude of belief is more extreme than would be observed by chance.'' Then, in the context of the full dataset, the magnitude annotation $m$ implicitly defines the ground-truth probability of our outcome:
\begin{equation}
    p = \mathbf{P}\{ m > M \}
\end{equation}
where $M$ is sampled from all dialogue annotations, and thus, $p$ can be computed from data. A separate agent (e.g., another interlocutor $\mathrm{B}$) can annotate their own perception $m'$ of $\mathrm{A}$'s belief, which is an expression of uncertainty/intensity of their belief about $\mathrm{A}$'s belief. This can then be calibrated to a probability estimate about the outcome $\gamma$, using the same formula:
\begin{equation}
    P = \mathbf{P}\{ m' > M \}.
\end{equation}
The important qualities of this outcome formulation are that: (a) it implicitly defines both uncertainty annotations and (ground-truth) calibration functions, which are not available in typical forecasting problems; and (b) it is general, since it implicitly models certainty about any belief for which we have magnitude annotations. 

The semantics of the outcome are, in fact, not much different than a more typical decision ``whether $\mathrm{A}$ believes \_\_\_\_\_'' instead asking a question about relativity of belief, to induce the (implicit) certainty annotations from those (magnitude annotations) that already exist. As a caveat, this outcome format does not work for calibrating uncertainty/intensity of beliefs about many types of ``future events''; e.g., whether a deal will occur. In these contexts, the human expression may need to be calibrated with data in order to map human expressions of belief intensity to the same scale as ground-truth outcome probabilities for comparison.
\subsection{Prompts}
\label{sec:prompts}
We use a common system prompt for all models: 

\textit{You are TheoryOfMindGPT, an expert language model at using your theory-of-mind capabilities to predict the beliefs and actions of others in human conversations. You will be given a potentially unfinished conversation between two speakers. Put yourself in the mindset of the speakers and try to reason about the requested conversation outcome. Use the keyword "CERTAINTY" to report your prediction for the outcome of interest. Report your answer on a scale from 1 to 10 with 1 indicating "not likely at all" and 10 indicating "almost certainly". For example, "CERTAINTY = 7".}

The user-role prompt is also common, with slight variations by corpora, or inclusion of demographics. Here is an example for MultiWOZ.

\textit{In the following conversation segment, a human user is interacting with an AI task assistant. **\texttt{insert conversation, set off by white space}**. Now, fast-forward to the end of the conversation. How certain is the user that they (the user) are more satisfied than would occur by chance? Let's think step by step, but keep your answer concise (less than 100 words).}

We found that ``Let's think step by step'' increased the rate of an explanation associated with the answer. This agrees with the findings of \citet{kojima2022large}. We also found that ``keep your answer concise (less than 100 words)'' was important to prevent models from going on too long without providing an answer (more than 256 tokens).
\begin{table}[]
\small
\begin{tabular}{llllll}
         & \texttt{Llama3} & \texttt{Mixtral} & \texttt{Gemma} & \texttt{GPT} & \texttt{Avg}   \\\hline
greedy & 0.5        & 2.2      & 2.2      & -0.7    & 1.0  \\
BoT      & \textbf{4.8}        & \textbf{2.6}      & \textbf{4.4}      & \textbf{3.3}     & \textbf{3.8}
\end{tabular}
\caption{Comparison of greedy sampling (temperature = 0) and BoT, illustrating distinction between them. Results show explained variance on MultiWOZ (1TUQ). BoT beats greedy sampling for all models. As final model prediction is conditioned on preceding explanation, setting temperature to 0 does not provide an adequate summary of the true (intractable) sampling distribution. BoT, on the other hand, approximates the mean of the true sampling distribution and provably lowers the variance of the model inference.}
\label{tab:mce_v_temp}
\end{table}
\subsection{Optimization}
\label{sec:optim}
We used \texttt{scikit-learn} to implement all scaling and fine-tuning algorithms \citep{scikit-learn}. Ordinary least squares is used to optimize both scaling methods, while SGD is used for the linear fine-tuning method. The neural regression head has a single hidden layer of dimension 100. The random forest has 100 trees of maximum depth 5. All other optimization parameters (e.g., for regularization) are the defaults of the library. 
\subsection{Model Comparison}
Table~\ref{tab:tuq-1-datasets} reports explained variance for 1TUQ for \texttt{DF LS}, split by model and dataset. \texttt{GPT-4o} and \texttt{Mix 8x22B} offer the best performance with \texttt{GPT} beating out \texttt{Mixtral} for first place, primarily on MultiWOZ. Interestingly, \texttt{Gemma 7B} (with BoT) outperforms two of the larger models, on average. The performance of \texttt{Llama3} \texttt{70B} was also surprising, as it is often improved by much smaller models (if they use BoT). Table~\ref{tab:tuq-2-ablation} also reports explained variance, split by model, for 2TUQ. Here, \texttt{Gemma} does not perform as well and neither does \texttt{GPT-4o} (i.e., the fair comparison across tables is \textit{without} demographic data). The most successful models are the \texttt{Mixtral} models, suggesting a unique advantage from their training data (closed-source) or their MoE architectures. \texttt{GPT 3.5} also shows promise in 2TUQ, under one setting (BoT, no demographics), but is less robust to perturbations among settings.
\subsection{Case Study: Does Considering Uncertainty Improve ToM Predictions Outright}
\label{sec:case_study}
\begin{table}
\centering
\begin{tabular}{@{}lllll@{}}
\toprule
            & \multicolumn{2}{l}{ACC}                           & \multicolumn{2}{l}{F1}     \\                       \cmidrule(lr){2-3}\cmidrule(lr){4-5}
uncertainty & \xmark & \checkmark & \xmark & \checkmark \\ \midrule
Llama 3 8B  & 60.5                  & 65.5                      & 71.7                  & 77.8                      \\
Llama 3 70B & 68.5                  & 70.5                      & 80.0                  & 82.2                     \\ \bottomrule
\end{tabular}
\caption{Accuracy and F1 of Llama 3 series models on \textbf{CaSiNo} corpora. We consider a binary conversation outcome -- whether both speakers are satisfied with the negotiation -- as suggested by \citet{sicilia2024deal}, to test our uncertainty estimates on a simple ToM belief prediction task. Results are shown with and without use of uncertainty estimation (\S~\ref{sec:belief_pred}) to make the prediction. These results show how reasoning about others' uncertainty can help language models, even in more traditional ToM tasks.}
\label{tab:belief_pred}
\end{table}
Throughout the paper, we have argued for the importance of estimating the uncertainty in others' beliefs, pointing to substantial existing literature as well as a few motivating examples. Here, we show how reasoning about others' uncertainty can even help language models to improve their accuracy at a traditional ToM task (i.e., an existing belief prediction task). Specifically, we use a belief prediction task built on one of the experimental corpora from \S~\ref{sec:exp}: CaSiNo. In this campsite negotiation corpora, annotations for satisfaction are provided on a Likert scale, but have clear semantic descriptions (e.g., ``very satisfied''), making it easy to create a binary labeling scheme for the outcome ``speakers are satisfied'' or not. We use the scheme outlined by \citet{sicilia2024deal} in their conversation forecasting work, where a conversation is labeled 1 if both speakers are satisfied  and 0 otherwise. We ask the model to make this prediction in two ways: (1) with an estimate of the speakers' uncertainty about this outcome and (2) without this estimate of uncertainty, making a simple binary prediction. For the uncertainty estimate, all answers greater than 5 are mapped to a prediction of 1 (i.e., a prediction that both users are satisfied). Overall, we use a largely similar prompt as shown in \S~\ref{sec:prompts} and used in our previous experiments. Results in Table~\ref{tab:belief_pred} are promising, showing that a language model's inferences can be more accurate when they reason about the uncertainty of others' beliefs to make predictions (rather than making a binary choice). This is especially true for the smaller Llama 3 model. These results are in line with our central argument that considering uncertainty in ToM is an important skill to evaluate. 

\end{document}